\documentclass[11pt, conference]{IAES2024}
\usepackage[top=1.5cm, bottom=1.5cm, left=2.4cm, right=2.4cm, includehead, includefoot, heightrounded]{geometry}

\usepackage{hyperref}
\hypersetup{pdftex,colorlinks=true,allcolors=blue}
\usepackage{hypcap}

\let\labelindent\relax

\usepackage{booktabs}
\usepackage[shortlabels]{enumitem}
\usepackage{newclude}
\usepackage{blindtext}
\usepackage{mathptmx}
\usepackage[english]{babel}
\usepackage{amsthm}

\theoremstyle{definition}

\theoremstyle{plain}
\newtheorem{theorem}{Theorem}[section]
\newtheorem{corollary}{Corollary}[theorem]
\newtheorem{lemma}[theorem]{Lemma}
\theoremstyle{remark}

\usepackage{wrapfig}
\usepackage{titlesec}
\titleformat{\section}
{\normalfont\rmfamily\Large\bfseries}
  {\thesection}{1em}{}

 \titleformat{\subsection}
{\normalfont\rmfamily\bfseries}
  {\thesubsection}{1em}{}
\titlespacing*{\section} {0pt}{2.5ex}{0.6ex}
\titlespacing*{\subsection} {0pt}{1.5ex}{0.6ex}
\usepackage{fancyhdr} 
\usepackage{graphicx} 
\usepackage[textfont=bf, font = normalsize]{caption} 
\usepackage{floatrow} 
\usepackage{amssymb}
\usepackage{amsfonts}
\usepackage{amsmath}
\usepackage[square,numbers,sort&compress]{natbib}
\floatstyle{plaintop} 
\restylefloat{table} 
\usepackage{subfigure} 
\renewcommand{\headrulewidth}{0pt}

\makeatletter
\renewcommand{\thesection}{\arabic{section}}
\renewcommand{\thesubsection}{\thesection.\arabic{subsection}}
\renewcommand{\p@subsection}{}
\renewcommand{\thesubsubsection}{\thesubsection.\arabic{subsubsection}}
\renewcommand{\p@subsubsection}{}
\makeatother
\usepackage{xcolor}
\begin{document}

\newcommand{\bing}[1]{{\color{blue}{\small\bf\sf [bing: #1]}}}
\newcommand{\changnan}[1]{{\color{red}{\small\bf\sf [changnan: #1]}}}

\title{\LARGE{Conditions for Length Generalization in Learning Reasoning Skills}


\vspace{-8mm}
}
\author{\textbf{Changnan Xiao\textsuperscript{1} and Bing Liu\textsuperscript{2} 
}
\vspace{+2mm}
\\
\textsuperscript{1}\small{ChangnXX.github.io}\\
\textsuperscript{2} \small{Department of Computer Science, University of Illinois Chicago}\\
\small{changnanxiao@gmail.com, liub@uic.edu}}
\maketitle
\thispagestyle{fancy} 
\pagestyle{plain}
\vspace{7mm}
\begin{abstract}
\noindent
Reasoning is a fundamental capability of AI agents. Recently, large language models (LLMs) have shown remarkable abilities to perform reasoning tasks. However, numerous evaluations of the reasoning capabilities of LLMs have also showed some limitations. An outstanding limitation is \textit{length generalization}, meaning that when trained on reasoning problems of smaller lengths or sizes, the resulting models struggle with problems of larger sizes or lengths. This potentially indicates some theoretical limitations of generalization in learning reasoning skills. These evaluations and their observations motivated us to perform a theoretical study of the length generalization problem. This work focuses on reasoning tasks that can be formulated as Markov dynamic processes (MDPs) and/or directed acyclic graphs (DAGs). It identifies and proves conditions that decide whether the length generalization problem can be solved or not for a reasoning task in a particular representation. Experiments are also conducted to verify the theoretical results.
\end{abstract}

\normalsize
\vspace{-0mm}
\section{Introduction}
\label{sec-intro}
Reasoning is an important part of Artificial Intelligence (AI) as it is a necessary capability of an intelligent agent. In recent years, learning to reason has been actively studied in the machine learning and natural language processing  communities~\citep{yu2023nature,storks2019commonsense,yang2023logical,helwe2021reasoning}. Recent \textit{large language models} (LLMs) like ChatGPT~\citep{openai2022chatgpt} and GPT4~\citep{openai2023gpt4} have been shown to perform sophisticated reasoning tasks in very impressive manners~\citep{brown2020language,suzgun2022challenging,saparov2022language,liu2023evaluating,xu2023large}. 

This success also invited many researchers to conduct systematic evaluations of the reasoning capabilities of LLMs.  
These evaluations showed impressive results on many reasoning problems, but at the same time  also revealed some limitations~\cite{liu2023evaluating,xu2023large}. {For example, several researchers showed that LLMs 
often encounter difficulties on simple multiplication, division, and addition of large numbers~\citep{nogueira2021investigating,qian2022limitations}.} 
 \cite{arkoudas2023gpt} listed a diverse set of reasoning tasks that GPT4 struggles with, which led the author to argue that most of the reasoning tasks successfully performed by GPT4 might already be in the pre-training dataset or very similar to some problems in the pre-training dataset. Evaluations conducted on several latest LLMs also showed that they struggled with many reasoning tasks \cite{gendron2023large,tang2023large}. 

The popular solution for improving the reasoning capabilities of LLMs is to use the approach called \textit{Chain of Thought} (\textbf{CoT})~\cite{wei2022chain} or \textit{Scratchpad}~\citep{nye2021show}. The CoT idea is natural. It simply adds the \textit{intermediate steps} for each reasoning problem in the training data. 
For example, the training sample for calculating $2+3 \times 4$ may be presented as $2+3 \times 4 = 2+12 = 14$. Several researchers have used CoT to improve the accuracy of arithmetic operations with good success \cite{wei2022chain,liu2023goat,lee2023teaching}. 
However, 
\cite{dziri2023faith} demonstrated experimentally that even with the detailed reasoning steps given in the training data, 
the learned models still fail to learn the operations in a generalizable manner for several reasoning problems. Specifically, the authors showed that when trained with smaller problems for a reasoning task (e.g., multiplication of two numbers), the model cannot generalize to larger problems. Several other authors have also noticed the problem, which is called the \textbf{\textit{length generalization}} problem~\cite{anil2022exploring,zhang2022unveiling}.   

The paper investigates the length generalization problem. It identifies and proves conditions that decide whether the length generalization problem can be solved or not for a reasoning task in a particular representation. Empirical evidences  are also given to verify the theoretical results (Sec. \ref{sec:experiment}).  

\section{Overview of Our Results}
\label{sec.overview}

This work focuses on studying the length generalization problem for reasoning tasks that can be formulated as \textit{Markov dynamic processes} (MDP) and/or \textit{directed acyclic graphs} (DAGs) structures, where MDP is a special case of DAG.
We progressively show: (1) given the MDP/DAG structure of a reasoning task, the condition for learning the \textit{causal function} (which performs a single reasoning step), (2) given the MDP/DAG structure and a well-learned causal function, a \textit{recursive formula} being able to solve the length generalization problem, 
and (3) given only unstructured data, the condition for learning the structure (i.e., predicting the elements involved in the next reasoning step) and solving the length generalization problem.

To illustrate our notations, we take the problem of calculating $2 + 3 \times 4$ as an example. A causal 
function infers one step in the reasoning/calculation process {(e.g., as specified in CoT).} In this example, $3 \times 4 = 12$ and $2 + 12 = 14$ are both one casual/reasoning step. 
In the arithmetic calculation on natural numbers $\mathbb{N}$, the causal function is $f: \mathbf{X} \rightarrow \mathbb{N}$, where $\mathbf{X} = \mathbb{N} \times \{+, -, \times, /\} \times \mathbb{N}$ is the input space of the causal function.

\begin{wrapfigure}[10]{r}{2.8in}
\vspace{-4mm}
\caption{
An example of MDP and DAG.
} 
\includegraphics[width=0.9\linewidth]{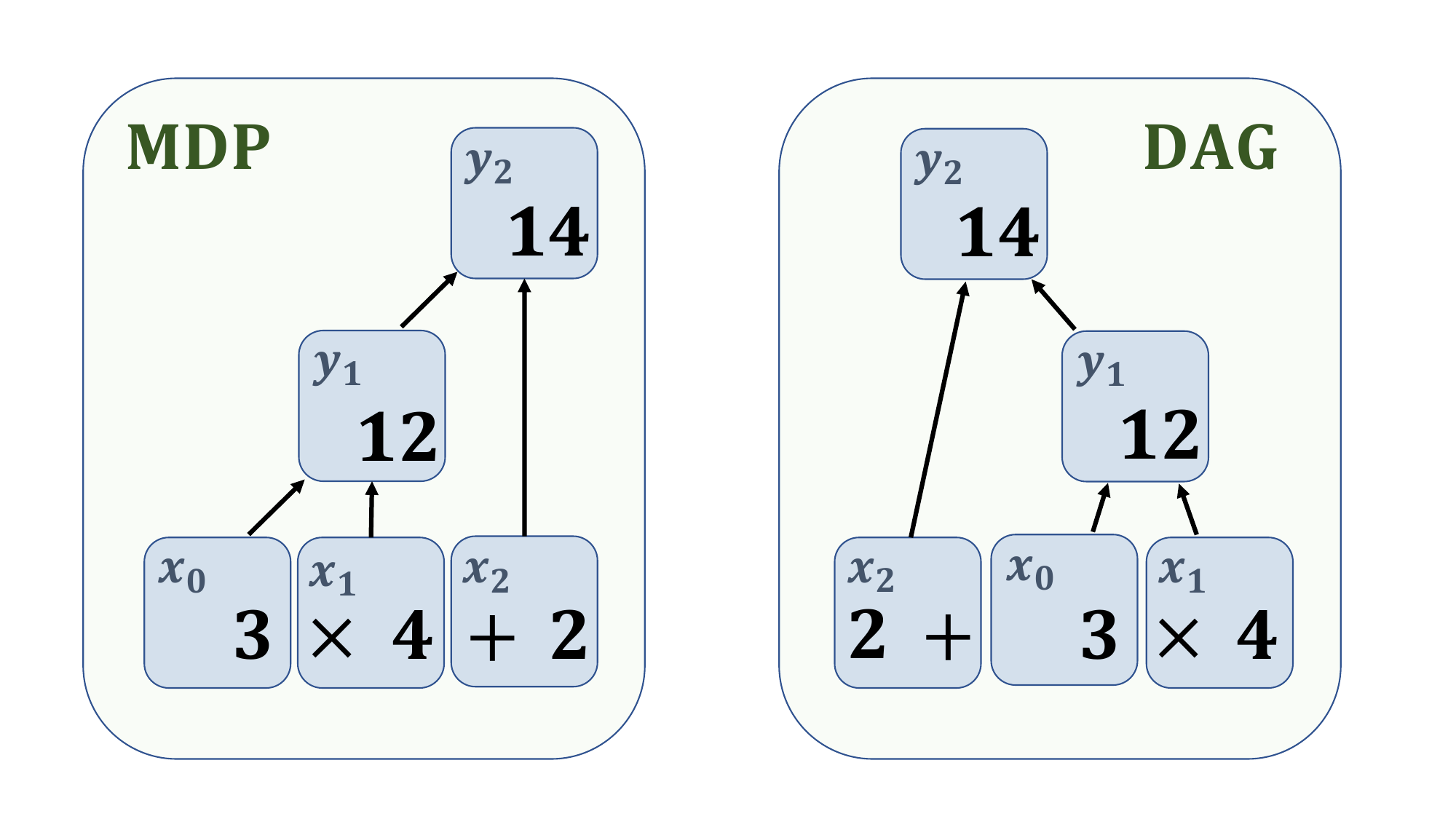}
\label{fig:example_mdp_dag}
\end{wrapfigure}

We define a dynamic process 
as a process that maps a sequence of elements to a sequence of elements from left to right 
representing a reasoning process.
{For instance, $3 \times 4 + 2$ defines a dynamic process with $x_0 = 3, x_1 = (\times, 4), x_2 = (+, 2), y_1 = f(x_0, x_1) = 3 \times 4 = 12, y_2 = f(y_1, x_2) = 12 + 2 = 14$,} where $f$ is the dynamic/causal function. Since $f$ only depends on the latest $x$ and $y$, this dynamic process is called a Markov dynamic process (MDP) (see Figure~\ref{fig:example_mdp_dag}), which reasons only sequentially and it is very limiting.
DAG is a generalization of MDP and is much more powerful.\footnote{~Some prior empirical works have used DAGs to represent reasoning problems \citep{shao2022chaining,cao2021bottom,huang2022directed}. Their goal was to generate DAGs from natural language descriptions of the reasoning problems. We also note that some reasoning problems cannot be modeled as DAGs, e.g., temporal and spatial reasoning problems.} 
{For instance, the reasoning process of $2 + 3 \times 4$ can be represented as the DAG in Figure~\ref{fig:example_mdp_dag}. }
The topological ordering of the DAG gives $x_0 = 3, x_1 = (\times, 4), x_2 = (2, +), y_1 = f(x_0, x_1) = 3 \times 4 = 12, y_2 = f(x_2, y_1) = 2 + 12 = 14$, where $f$ is the causal function. For simplicity, we will always use \textit{causal function} and it also means \textit{dynamic function} for dynamic process. 

We define a \textit{recursive formula} as a function that recursively applies the causal function step-by-step to solve a reasoning problem according to its structure, e.g., $2 + 3 \times 4 = 2 + 12 = 14$ (see the DAG image in Figure~\ref{fig:example_mdp_dag}). In Secs \ref{sec:dp} and \ref{sec: dag}, the structure (MDP/DAG with calculation steps) is given. 
In Sec. \ref{sec:cot_and_others}, the structure is unknown and needs to be learned, which is the practical scenario.  

In Sec.~\ref{sec:dp} and Sec.~\ref{sec: dag}, we start from a simple or ideal scenario where the structure of the reasoning problem is given as an MDP or a DAG. We prove the following: 
\begin{itemize}
\item the causal function can be perfectly learned (called \textit{well-learned}) only when the input space of the causal function is finite, $|\mathbf{X}| < \infty$. 
\item when evaluating after training, given the MDP/DAG of the reasoning problem, solving the problem by recursively applying a well-learned causal function can generalize from smaller training problem/sample sizes to testing of larger problem sizes.
\end{itemize}
In Sec.~\ref{sec:cot_and_others}, we study the realistic usage scenario where the structure of the problem is unknown and only unstructured sequence data with CoT is given in training, e.g., a sequence of elements like $2+3 \times 4 = 2+12=14$. We will not study the case when the CoT (or the intermediate reasoning) steps are not given but only the input and the output like $2+3 \times 4 = 14$ as several researchers~\cite{wies2023sub,feng2023towards,malach2023auto} have shown that this case is not learnable. We focus on the following,
\begin{itemize} 
\item learning to predict which elements should be the input to the well-learned causal function to calculate or reason next, i.e., learning the ordering of calculations. We propose an important notion $R$, the \textit{maximal input element distance for a causal/reasoning step}. For instance, for the arithmetic expression $2+3 \times 4$ without using `(' and `)', we have $R=2$ because the elements that should be calculated next are in a window of length $3$, e.g., $3 \times 4$, where the maximal input elements distance is $2$ (the distance between $3$ and $4$). If the arithmetic calculation contains `(' and `)', e.g., $2 + (3 \times 4$), then $R = 4$ because the elements that should be calculated next is in a window of at most of length $5$, e.g., $( 3 \times 4 )$.
\item proving that the condition for (length) generalization from smaller/shorter training problems to larger/longer test problems is $R < \infty$. 
\end{itemize}

As discussed earlier, several papers have reported experiments showing that deep learning models struggle with length generalization for some reasoning problems even with the help of CoT or Scratchpad \cite{anil2022exploring,dziri2023faith,zhang2022unveiling}. Based on our theory, the CoT formulations of these problems suffer from either $|\mathbf{X}| = \infty$ or $R = \infty$, which causes the length generalization issue. In Sec.~\ref{sec:experiment}, we will see that the same reasoning problem may be solved if a different CoT representation is used, which satisfies $R < \infty$. 

{Learnability studies of reasoning problems using CoT have been reported in~\cite{wies2023sub,feng2023towards,malach2023auto} for neural networks. However, these studies all under the given problem length/size $N$. Their statements are like ``for $\forall\, N > 0$, given a dataset with training problems no longer than $N$, for any problem with length $N' \leq N$, it can be solved under a PAC upper bound.'' The key limitation is that training problem length and testing problem length are the same, or the upper bound depends on the testing problem length. Our theory doesn't have this limitation and works on a more general scenario with training on smaller length problems and testing on larger length problems. Our statement is that ``for $\forall\, N > 0$, given a training dataset with problems no longer than $N$, problems of any length $N'$ can be solved if $|\mathbf{X}| < \infty$ and $R < \infty$.''} {Our work can be seen as complementary to theirs in the {context of neural networks (as our results are independent of specific learning paradigm or algorithm).} Their learnability results still apply. We add conditions under which the learned function can extrapolate to larger lengths than $N$ to achieve length generalization.}

Length generalization is related to \textit{out-of-distribution} (OOD) generalization in~\cite{abbe2023generalization}. However, there is a major difference. In length generalization, since the training data in practice is always finite, the maximal size or length of the training problems is also finite, but a larger size/length problem can always appear in testing, which can be seen as OOD, but such an OOD is unavoidable regardless how much data is used in training as long as it is finite. 
{In \cite{abbe2023generalization}, OOD is defined as some subspace (i.e., some values or value combinations) never appeared in training but appeared in testing. 
This type of OOD generalization can be solved with more and/or diverse training data.} We do not study this problem. Further, the study in \cite{abbe2023generalization} was  under some specific model architectures and activation functions. Our work is algorithm and architecture independent and we focus on length generalization. 


In summary, to learn to solve reasoning problems and to overcome the {length generalization} issue, we propose \textit{three sufficient conditions}, \textbf{(i)} the input space $\mathbf{X}$ of a causal/reasoning step is finite, \textbf{(ii)} the problem is solved recursively based on CoT,
and \textbf{(iii)} the \textit{maximal input element distance} $R$ of the unstructured representation for a causal/reasoning step is finite. 

\section{Our Results}
\subsection{Markov Dynamic Process (MDP)}
\label{sec:dp}




This section considers sequential reasoning problems. In such a problem, the inference steps are made in a single sequence. 
We take the arithmetic problem in a prime field $F_p = \mathbf{Z} / p\mathbf{Z}$ as an example, which can be formulated into a sequential reasoning problem. For instance, we solve $((((a_0 + a_1) \times a_2) - a_3) / a_4)$ sequentially by $b_1 = f(a_0, +, a_1)$, $b_2 = f(b_1, \times, a_2)$, $b_3 = f(b_2, -, a_3)$, and $b_4 = f(b_3, /, a_4)$, where $f$ is the definition of $(+, -, \times, /)$ in $F_p$. 
In this section and the next, we do not regard `(' and `)' as basic elements in the arithmetic problem. When they show up, they only indicate the order of the calculation steps. In Sec. \ref{sec:cot_and_others}, we'll revisit them and they are significant when finding the elements to be calculated next. 

Denote $X$ to be the domain. $X$ contains all the input elements of all reasoning steps. 
Denote $Y$ to be the range. $Y$ contains all the output elements of all reasoning steps. 
$X$ and $Y$ can have multiple dimensions (see the 3-line addition example in Section~\ref{sec:experiment}).
For the arithmetic problem in $F_p$, we define $Y = F_p$ and $X = \{+, -, \times, /\} \times F_p$, since the arithmetic symbols or operators are also the input elements for each calculation step. 
For simplicity, we denote a sequence of elements in $X$ as 
\begin{equation}
x_{[i: j]} = (x_i, x_{i + 1}, \dots, x_j).
\end{equation}
Let $f: Y \times X^\infty \rightarrow Y$ be a  \textbf{casual function} (we can also call it a \textit{dynamic function}\footnote{~Note that \textit{dynamic function} here is not related to the term \textit{dynamic function} used in programming languages.}), which performs one reasoning step. 
We denote the dynamic process $\tau_f$ defined by $f$ as 
\begin{equation}
\tau_f( x_{[0: \infty]}) = (y_1, y_2, \dots),\ \text{where}\  \left\{
\begin{aligned}
    &y_1 = f(y_0, x_0), \\
    &\dots \\
    &y_{n + 1} = f(y_n, x_{[0: n]}). \\
\end{aligned}
\right.
\label{eq: def_dp}
\end{equation}
Without loss of generality, we always assume $y_0 = 0$. 

For instance, in the arithmetic expression $((((a_0 + a_1) \times a_2) - a_3) / a_4)$, $x_{[1: 4]} = ((+, a_1), \dots, (/, a_4))$,  where `(' and `)' only indicate the correct calculation order, 
the causal function $f: F_p \times \{+, -, \times, /\} \times F_p \rightarrow F_p$ defines the arithmetic calculations in $F_p$. 
And $\tau_f(x_{[0: 4]}) = (b_1, b_2, b_3, b_4)$, where $b_1 = f(a_0, +, a_1)$, $b_2 = f(b_1, \times, a_2)$, $b_3 = f(b_2, -, a_3)$, and $b_4 = f(b_3, /, a_4)$. 
The first calculation step $b_1 = f(a_0, +, a_1)$ corresponds to $y_1 = f(y_0, x_1)$, where $y_1 = b_1$, $y_0 = a_0$ and $x_1 = (+, a_1)$.

We say $\tau_f$ is a Markov dynamic process (MDP) if $y_{n + 1} = f(y_n, x_{[0, n]})$ depends only on $(y_n, x_n)$ for $\forall\, n > 0$. We denote $y_{n+1} = f(y_n, x_n)$, where $f: Y \times X \rightarrow Y$ is the causal function for MDP. The \textbf{input space} of the causal function is $\mathbf{X} = Y \times X$. We emphasize that the definition of the MDP only  requires that $f$ depends on the current $x_n$ rather than the past $x_i$'s. 

To understand the learning of MDP, we first study the causal function. 
Denote the training dataset of $f$ as $D$, which is a subset of the input space,  
\begin{equation}
D \subseteq \{(f(y, x), (y, x)) | (y, x) \in Y \times X\}.
\end{equation} For simplicity, we denote $D \subseteq \mathbf{X} = Y \times X$. 
For instance, in the arithmetic problem in $F_p$, we have $D \subseteq \{(f(a_1, *, a_2), a_1, *, a_2 )|\,a_1, a_2 \in F_p, * \in \{+, -, \times, /\}\}$. 
We assume  $|D| < \infty$. 

\begin{theorem}
\label{thm: full_dynamic_function}
For $|X|, |Y| < \infty$, i.e. $|\mathbf{X}| < \infty$, if $D = \mathbf{X}$, 
then there exists an approximation function $\hat{f}: Y \times X \rightarrow Y$ s.t. $\Hat{f}(y, x) = f(y, x),\, \forall\, (y, x) \in \mathbf{X}$. 
\end{theorem}
The proof is given in \textit{Appendix}~\ref{sec.dynamic.causal}. 
This theorem says that when $|X|, |Y| < \infty$, which implies $|\mathbf{X}| < \infty$, there exists an approximation function that perfectly approximates the causal function $f$ defined on $Y \times X$. 
For instance, when $D = F_p \times \{+, -, \times, /\} \times F_p$, there must exist a function $\Hat{f}$ that learns the same arithmetic rules as $f$'s in $F_p$. 
When $D \neq \mathbf{X}$, we have the following Corollary \ref{thm: not_full_dynamic_function} that shows the \textbf{impossibility} of perfectly approximating the causal function $f$. 

\begin{corollary}
\label{thm: not_full_dynamic_function}
For $|X|, |Y| < \infty$, i.e. $|\mathbf{X}| < \infty$, if $|Y| > 1$ and $D \neq \mathbf{X}$, then there exists an approximation function $\hat{f}: Y \times X \rightarrow Y$ s.t. $\Hat{f}(y, x) = f(y, x),\,\forall\, (y, x) \in D$ and $\Hat{f}(y, x) \neq f(y, x),\, \forall\, (y, x) \in \mathbf{X} \backslash D.$
\end{corollary}

The proof is given in \textit{Appendix}~\ref{sec.dynamic.causal}. Corollary \ref{thm: not_full_dynamic_function} shows that when $D \neq \mathbf{X}$, there always exists a function that predicts correctly on $D$ but wrongly on the complement ($\mathbf{X} \backslash D$). 
\cite{abbe2023generalization} showed that the learned random feature model tends to learn an low-degree approximation function theoretically, which means that the generalization on the unseen is disappointing unless there exist some subtle underlying connections between the unseen and the hypothesis space. 
Thus, it supports the {\textit{\textbf{impossibility}} of learning the causal function when unseen (out-of-distribution) exist in testing.} Note that Corollary \ref{thm: not_full_dynamic_function} is independent of specific learning algorithms or hypothesis spaces. 
The conclusion can naturally be extended to $|X| = \infty$. 



\begin{corollary}
\label{thm: not_full_dynamic_function_inf_x}
For $|X| = \infty$, i.e. $|\mathbf{X}| = \infty$, if $|Y| > 1$, for $\forall\, m > 0$, there exists an approximation function $\hat{f}: Y \times X \rightarrow Y$ s.t. $\Hat{f}(y, x) = f(y, x),\,\forall\, (y, x) \in D$ and $| \{(y, x) | \Hat{f}(y, x) \neq f(y, x)\} | > m.$
\end{corollary}
The proof is deferred to \textit{Appendix}~\ref{sec.dynamic.causal}. Now we know the causal function $f$ can be well-learned only when $|\mathbf{X}| < \infty$,
regardless of the learning algorithm and the hypothesis space. 
When we look back at the MDP, it can simply be \textit{recursively solved} 
with the well-learned causal function. 
\begin{theorem}
\label{thm: recursive_markov_dp_solved}
For $|X|, |Y| < \infty$, i.e. $|\mathbf{X}| < \infty$, if $D = \mathbf{X}$, then there exists an approximation function $\Hat{f}: Y \times X \rightarrow Y$, the MDP $\tau_f$ can be recursively solved, i.e. $\forall\, n > 0,\,\tau_{\Hat{f}}( x_{[0: n]}) = \tau_{f}( x_{[0: n]})$.
\end{theorem}
The proof is given in \textit{Appendix}~\ref{sec.recursive.direct}. The theorem also guarantees that a MDP of \textit{any length} can be recursively solved.

\subsection{Directed Acyclic Graph (DAG)}
\label{sec: dag}

Now we know the MDP of an arbitrary length can be solved recursively by Theorem \ref{thm: recursive_markov_dp_solved}.
To generalize the result, we introduce the Direct Acyclic Graph (DAG), a directed graph with no directed cycles. 
In this section, we consider reasoning problems that can be structured into DAGs.
A topological ordering can order the vertices of a DAG, where every edge leaves a front vertex and enters a latter vertex. 
The existence of the topological ordering is an equivalent definition of DAG, i.e. every DAG has at least one topological ordering \citep{bang2008digraphs}. 
Therefore, every DAG can be transformed into a dynamic process. However, the dynamic process from a DAG may not be Markov. 
For instance, the causal function $f(y_n, x_{[0: n]}) = y_n + x_n + \dots + x_0$ defines a DAG as well as a dynamic process that always depends on all the previous $x_i$'s, which is not Markov. 
Even though a DAG is not a MDP, it's still possible to solve the DAG recursively. 

Denote a DAG as $G = (V, E)$, where $V$ is the set of all vertices and $E$ is the set of all edges. 
For easy understanding, we use $|G|$ $(= |V|)$ to represent the number of vertices in $G$. 
Denote $u \overset{e}{\rightarrow} v$ as when $v$ is reachable directly from $u$ by the edge $e$. 
Denote $s(v) = \{u \in V |\, \exists\, e \in E, u \overset{e}{\rightarrow} v\}$ to be all vertices (immediate in-neighbors) that can reach $v$ directly. 
Denote the \textit{in-degree} of each vertex $v$ as $|s(v)|$. 

For a reasoning problem structured as a DAG, when $u \overset{e}{\rightarrow} v$, we say that $u$ is the \textbf{input vertex} and $v$ is the \textbf{causal vertex}. 
Since a causal vertex can also be the input vertex of some other vertices, we do not distinguish the domain and the range as in Sec.~\ref{sec:dp}. 
With a slight abuse of notation, we also use $v$ to represent the value of vertex $v$. 
We simply denote $X$ to be both the domain and the range of all vertices. 
Let $f: X^{\sup |s(v)|} \rightarrow X$ to be the \textbf{causal function}, which is $v = f(s(v))$. The \textbf{input space} of $f: X^{\sup |s(v)|} \rightarrow X$ is $\mathbf{X} = X^{\sup |s(v)|}$. 
Note that $G$ defines the structure of the reasoning process and $f$ defines the values. 
For any $G = (V, E)$, let $V = \{v_1, \dots, v_{|G|}\}$ be a topological ordering of the vertices in $G$. 
Note that the vertices with $|s(v)| = 0$ are pure input vertices of the graph.
We say
\begin{equation}
G_f(\{v_i \,|\, i \leq |G|,\, |s(v_i)| = 0\}) = (v_1, \dots, v_{|G|}), \text{where}\ 
\left\{
\begin{aligned}
&v_1 = f(s(v_1)), \\
&\dots \\
&v_n = f(s(v_n)),
\end{aligned}
\right.
\label{eq: def_dag}
\end{equation}
is the induced graph defined by $f$. {The edge information is included in the topological ordering. } 

\begin{wrapfigure}[11]{r}{2.8in}
\vspace{-4mm}
\caption{
An example of notations. 
} 
\includegraphics[width=0.96\linewidth]{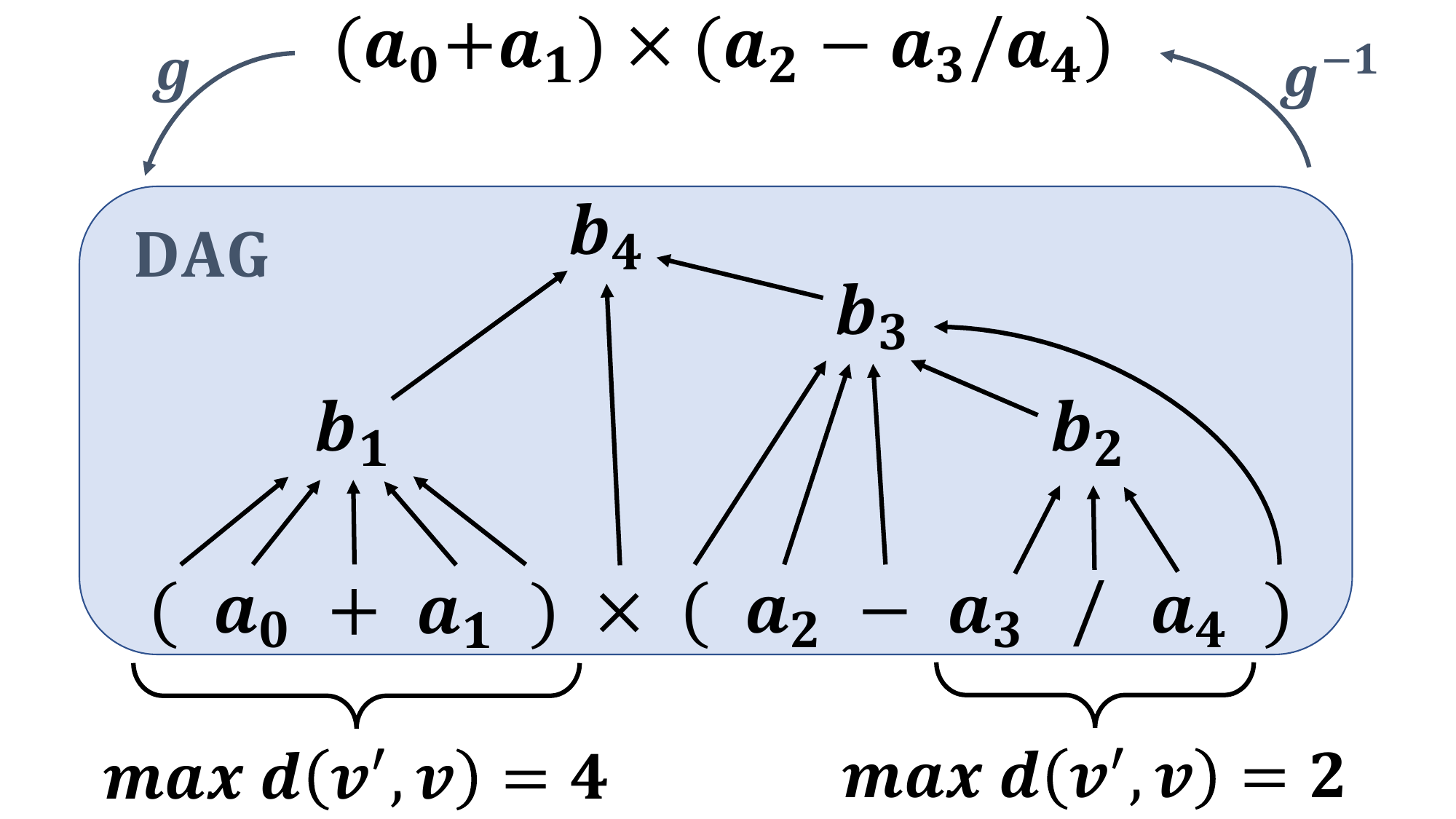}
\vspace{-3mm}
\label{fig:example_dag}
\end{wrapfigure}

The arithmetic problems in $F_p$ are DAGs. For instance, to calculate $(a_0 + a_1) \times (a_2 - (a_3 / a_4))$, we can use the following steps $b_1 = f(a_0, +, a_1)$, $b_2 = f(a_3, /, a_4)$, $b_3 = f(a_2, -, b_2)$, and $b_4 = f(b_1, +, b_3)$, which defines a DAG. The DAG defines the calculation steps and the causal function $f$ defines the arithmetic rules in $F_p$. When $f$ changes, the induced values on $G$ change. That's why we say $G_f$ is the induced graph of $G$ defined by $f$. 

We first study the property of $f$. 
Denote the training dataset of $f$ as $D$, which is a subset of the input space as 
\begin{equation}
D \subseteq \{(f(s(v)), s(v)) |\, v \in V\}.
\end{equation}
$D$ contains vertices from any $G$. For simplicity, denote $D \subseteq \mathbf{X} = X^{\sup |s(v)|}$ and $f(\mathbf{X}) = \{f(s(v))|\, s(v) \in \mathbf{X}\}$. We assume $|D| < \infty$, which is the case for any training dataset. 

\begin{theorem}
\label{thm: full_causal_function}
For $|X| < \infty$ and $\sup |s(v)| < \infty$, i.e. $|\mathbf{X}| < \infty$, if $D = \mathbf{X}$, 
then there exists an approximation function  $\Hat{f}: X^{\sup |s(v)|} \rightarrow X$,  s.t. $\Hat{f}(s(v)) = f(s(v)),\,\forall\, s(v) \in \mathbf{X}$.
\end{theorem}

The proof is given in \textit{Appendix}~\ref{sec.dynamic.causal}. The only difference between Theorem \ref{thm: full_causal_function} and Theorem \ref{thm: full_dynamic_function} is the assumption $\sup |s(v)| < \infty$, which is similar to the Markov assumption. These assumptions make the domain of $f$ to have finite dimensions, which guarantees the feasibility of learning $f$ with $|D| < \infty$. 
Similar to the MDP, when $D \neq \mathbf{X}$, we have the following \textbf{negative} or \textbf{impossibility} corollaries (proofs are given in \textit{Appendix}~\ref{sec.dynamic.causal}).

\begin{corollary}
\label{thm: not_full_causal_function}
For $|X| < \infty$ and $\sup |s(v)| < \infty$, i.e. $|\mathbf{X}| < \infty$, if $|f(\mathbf{X})| > 1$ and $D \neq \mathbf{X}$, then there exists an approximation function $\Hat{f}: X^{\sup |s(v)|} \rightarrow X$ s.t. $\Hat{f}(s(v)) = f(s(v)),\,\forall\,s(v) \in D$ and $\Hat{f}(s(v)) \neq f(s(v)),\,\forall\,s(v) \in  \mathbf{X} \backslash D$. 
\end{corollary}

\begin{corollary}
\label{thm: not_full_causal_function_inf_x}
For $\max(|X|, \sup |s(v)|) = \infty$, i.e. $|\mathbf{X}| = \infty$, if $|f(\mathbf{X})| > 1$, for $\forall\, m > 0$, there exists an approximation function $\Hat{f}: X^{\sup |s(v)|} \rightarrow X$ s.t. $\Hat{f}(s(v)) = f(s(v)),\,\forall\,s(v) \in D$ and $|\{s(v) | \Hat{f}(s(v)) \neq f(s(v))\}| > m$. 
\end{corollary}

Similar to that in MDP, the causal function is guaranteed to be well-learned only when $|\mathbf{X}| < \infty$,
regardless of the learning algorithm and the hypothesis space.
Noticing the similarity between $\sup |s(v)| < \infty$ and the Markov property, it's natural to extend the conclusion of MDP to DAG. 

We now show that a DAG can only be solved \textit{recursively}. 

\begin{theorem}
\label{thm: resursive_dag_solved}
For $|X| < \infty$ and $\sup |s(v)| < \infty$, if $D = X^{\sup |s(v)|}$, then there exists an approximation function $\Hat{f}: X^{\sup |s(v)|} \rightarrow X$, the DAG can be recursively solved, i.e. $\forall\,G = (V, E)$, $G_{\Hat{f}} = G_f$. 
\end{theorem}

The proof is given in \textit{Appendix}~\ref{sec.dynamic.causal}. 
Theorems \ref{thm: recursive_markov_dp_solved} and \ref{thm: resursive_dag_solved} show that when the structure of the problem is given, and the inner causal function $f$ is well-learned, a problem of any length or size can be solved.

So far we have assumed that the MDP or the DAG structure is given for a reasoning task. But in practice, neither of them is given. The next section deals with this more realistic scenario. 

\subsection{Unstructured Sequence Data}
\label{sec:cot_and_others}



{Sec. \ref{sec:dp} and Sec. \ref{sec: dag} assume the structure of the reasoning problem is given as a MDP or DAG.  
This section assumes the structure is unknown.} We discuss the possibility to learn to transform an input sequence of unstructured elements into a structured DAG. We achieve this recursively by learning to find elements that should be reasoned next, like recursively doing topological ordering of the underlying DAG (we do not physically construct the DAG). For instance, $(a_0 + a_1) \times (a_2 - (a_3 / a_4))$ is a sequence of elements (including `(' and `)') and the ordering of calculation needs to be learned. Note that we will not use MDP subsequently anymore but only DAG as DAG is more general and powerful. 

Chain-of-Thought (CoT) \citep{wei2022chain} is a well-known approach to solving reasoning problems for unstructured data by giving the intermediate reasoning steps in the training data. We use $(a_0 + a_1) \times (a_2 - (a_3 / a_4))$ in $F_p$ as an example to see how CoT works and how it connects to DAG and recursive reasoning on DAG. The process of CoT is 
\begin{align}
(a_0 + a_1) \times (a_2 - (a_3 / a_4)) 
= b_1 \times (a_2 - b_2) 
= b_1 \times b_3 
= b_4.
\end{align}
CoT actually does three things or has three sub-steps in each reasoning/causal step. Sub-step 1 decides a part of the topological order of the underlying DAG, specifically, to find the vertices that will potentially be involved in the next reasoning step, i.e., vertices in $\{s(v) |\,v \in W^0\}$ ($W^0$ are vertices to be valued in the next reasoning step, which is defined later), before each reasoning step.
Sub-step 2 applies the causal function (previously learned), e.g., $b_1 = f(a_0, +, a_1)$, $b_2 = f(a_3, /, a_4)$, etc. 
Sub-step 3 puts the result back into the unstructured sequence. Since the arithmetic example decreases the number of elements at each step, sub-step 3 looks natural. However, it's possible to have a reasoning problem whose underlying DAG induces multiple vertices with multiple elements after one reasoning step. The representation by the unstructured data of the result after one reasoning step shouldn't be ambiguous; otherwise the reasoning problem itself is not well-defined. 

To define a well-defined reasoning problem, we need more notations. For a DAG $G = (V, E)$, we have defined $s(v) = \{u \in V |\, \exists\, e \in E, u \overset{e}{\rightarrow} v\}$ to be all vertices that can reach $v$ directly ($v$'s in-neighbors). Now we define $t(v) = \{w \in V |\, \exists\, e \in E, v \overset{e}{\rightarrow} w\}$ to be all vertices that can be reached from $v$ directly ($v$'s out-neighbors). We define $(s \circ t)(v) = \{v' \in V |\, \exists\, w \in V,\, \exists\, e_1, e_2 \in E, v \overset{e_1}{\rightarrow} w, v' \overset{e_2}{\rightarrow} w \}$ to be vertices that can reach any causal vertex in $t(v)$.

Now we define what a well-defined reasoning problem is in an unstructured sequence. Denote the initial unstructured data to be a sequence of elements $s^0 = s_1^0 s_2^0 \dots s_{i_0}^0$. Let $g$ be the function that transforms the unstructured data into a structured DAG. We write $G^0 = (V^0, E^0) = g(s^0)$. We say a vertex is \textit{valued} if a value is assigned to the vertex. In $G^0$, all vertices with $|s(v)| = 0$ have been valued as they are pure input vertices, but all causal vertices with $|s(v)| > 0$ haven't been valued. After one reasoning step on $G^0$, the vertices $W^0 = \{v \in G^0|\, s(v) \subseteq \{v \in G^0|\, |s(v)| = 0\}\}$ are valued. Then the vertices $U^0 = \{v \in \bigcup_{v \in W^0} s(v)|\, t(v) \subseteq W^0 \}$, whose causal vertices are valued, become useless in the following reasoning steps. We denote $G^1 = G^0 \setminus U^0$ to be the subgraph of $G^0$ after removing all vertices in $U^0$ and all edges from $U^0$. To represent $G^1$, let $g^{-1}$ be the inverse function that maps a DAG back to the unstructured data (which may not be unique). 
We write $s^1 = s_1^1 s_2^1 \dots s_{i_1}^1 = g^{-1} (G^1)$. 

We say the reasoning problem is \textit{well-defined} with the unstructured data if $g(g^{-1} (G)) = G,\,\forall\, G$. Note that $g(g^{-1} (G)) = G$ represents that the DAG is isomorphic after the composing transformation $(g \circ g^{-1})$. 


Now we describe the three sub-steps of CoT again, with a well-defined reasoning problem. Sub-step 1 is to find  the input vertices $\{s(v) |\,v \in W^0\}$ and the next valued vertices $W^0 = \{v \in G^0|\, s(v) \subseteq \{v \in G^0|\, |s(v)| = 0\}\}$.
Note that CoT only needs a one-step-forward subgraph instead of the entire graph. It's unnecessary to apply the original $g$. Instead, we define 
\begin{align}
\Tilde{g}(s^0) \overset{def}{=} \{s(v) |\, v \in g(s^0), s(v) \subseteq \{v \in g(s^0)|\, |s(v)| = 0\}\},
\end{align}
which finds the combinations of input vertices that infer causal vertices for one causal step. 
Different from $g$ which constructs the entire graph, $\Tilde{g}$ only needs to construct the graph for one causal/reasoning step. 

Sub-step 2 applies the causal function $f$ to infer 
\begin{equation}
f(\Tilde{g}(s^0)) \overset{def}{=} \{f(s(v)) | \ s(v) \in \Tilde{g}(s^0)\}.
\end{equation}

Sub-step 3 puts the resulting one-step-forward subgraph into the unstructured data. 
Note that $s_1^0,\dots, s_{i_0}^0$ are vertices of $g(s^0)$, where $v \in s^0$ means $v$ is a vertex of $g(s^0)$ that is valued by some $s_j^0 \in s^0$. 
Since the vertices $s^0$ and the one-step-forward vertices $f(\Tilde{g}(s^0))$ construct a subgraph of $g(s^0)$, and $g^{-1}$ is well-defined, we simply map the subgraph of vertices $(s^0, f(\Tilde{g}(s^0)))$ into the unstructured data by 
\begin{equation}
s^1 = g^{-1} (f(\Tilde{g}(s^0)), s^0).
\end{equation}
Sub-step 3 can be complex. For instance, there may exist input vertices that infer multiply causal vertices by $f$, i.e. $|f(s(v))| > 1$, then the question is how to represent $f(s(v))$ by the unstructured data. It's also a question where to put $f(s(v))$ back in $s^0$ when $s(v)$ is not a successive sub-sequence of $s^0$. The other question is how to deal with the useless vertices $U^0 = \{v \in \bigcup_{v \in W^0} s(v)|\, t(v) \subseteq W^0 \}$. 

In this work, we define $s^1 = g^{-1} (f(\Tilde{g}(s^0)), s^0)$ by specific rules of the reasoning problem. 
But it's still a question whether sub-step 3 can be learned in general. We leave this for the future study. 

In Sec. \ref{sec: dag}, we discussed the causal function, which is used in sub-step 2 of CoT. Now we discuss the possibility of learning sub-step 1. 
Denote 
\begin{equation}
D \subseteq \{(\Tilde{g}(s), s)|\, s = g^{-1}(G)\}
\end{equation}
to be the dataset of $\Tilde{g}$. $G$ represents the underlying structure of the reasoning problem, and $s = g^{-1}(G)$ is the corresponding unstructured or sequence data of $G$. 
When $v_i = s^0_i$ and $v_j = s^0_j$, denote  
\begin{equation}
d(v_i, v_j) \overset{def}{=} |i - j|,
\end{equation} which defines the distance between two vertices that were originally in $s^0$ to be their index distance in $s^0$. 

We now introduce the important notion of $R$, the \textit{maximal input element distance for a reasoning step}, which decides whether \textit{\textbf{length generation}} can be achieved. 
{\color{black} Define 
\begin{equation}
R \overset{def}{=} \sup \max_{v' \in (s \circ t)(v)} d(v', v). 
\end{equation}
The $\sup$ is for any element $v$ in the unstructured sequence of a problem, the $\max_{v' \in (s \circ t)(v)}$ is the maximal distance between any two elements that should involved in the next calculation/reasoning. 
$R$ is the maximal distance between the elements that should be calculated next in the same calculation step of the problem.}
\begin{theorem}
\label{thm: find_input_vertices_is_possible}
For $R < \infty$, if $D = X^{4R +1}$,
then there exists an approximation function $\Hat{g}: X^{4R + 1} \rightarrow 2^{4R + 1}$ 
s.t. $\Hat{g} (s) = \Tilde{g} (s),\,\forall\, s \in X^{4R + 1}$. 
\end{theorem}

The proof is given in \textit{Appendix}~\ref{sec.mied}. It says that for any problem with $G$ as the underlying DAG structure of the problem ($G$ is not accessible in this section), $\forall\, v \in g^{-1}(G),\,\forall \, v' \in (s \circ t)(v)$, $d(v', v)$ is uniformly bounded by a constant $R$. $R = \sup \max_{v' \in (s \circ t)(v)} d(v', v)$ is a primitive property of the reasoning DAG $G$ and function $g^{-1}$ transforms $G$ into the unstructured data.
For instance, if $s^0 = (a_0 + a_1) \times (a_2 - (a_3 / a_4))$, we have $\Tilde{g}(s^0) = \{(a_0 + a_1),\,(a_3 / a_4)\}$. It's easy to see $R = 4$ for arithmetic problems with `(' and `)'. 

Theorem \ref{thm: find_input_vertices_is_possible} requires $D = X^{4R + 1}$ because the combinations within a $(4R+1)$-length window {cover all cases that enable a correct order to be learned}. 
For instance, in `$\dots e \times a + b + c \times d \dots$', it's obvious that $R = 2$ because each operation or reasoning step involves 3 elements, e.g., $e \times a$, $a + b$, $b + c$, $c \times d$. To decide whether $b$ should be calculated first, we look at neighbors of $b$ in radius $R$, which is a window of length $2R + 1$. In this window $a + b + c$, we know $a + b$ should be calculated first. However, this is not enough because we also need to consider neighbors of $a$ in radius $R$ to validate whether $a + b$ should be calculated first. {In the window $e \times a + b$, we know $e \times a$ should be calculated first. Therefore, in this $(4R+1)$-length window $e \times a + b + c \times d$, we know $b$ shouldn't be calculated first.} A $(4R+1)$-length window guarantees correctness when an element belongs to multiple combinations of input vertices. $D = X^{M}$ ($M < 4R+1$) cannot guarantee the correctness and an example is given in Appendix \ref{sec.mied}.

When $R < \infty$, Theorem \ref{thm: find_input_vertices_is_possible} holds, where a learned function $\Hat{g}$ predicts which elements in a $(4R+1)$-length window should be calculated/reasoned next. When $|\mathbf{X}| < \infty$, Theorem \ref{thm: full_causal_function} 
holds, where a learned causal function $\Hat{f}$ takes the predicted elements of $\Hat{g}$ as input to predict the value of the next/causal vertex. By Theorem \ref{thm: resursive_dag_solved}, the recursive process solves the problem of arbitrary length, i.e., achieving \textbf{length generalization}.

When $R = \infty$, it means for any $N > 0$ in training, there exists an input element $v$ s.t. $\max_{v' \in (s \circ t)(v) } d(v', v) > N$ in testing, i.e. the distance between elements to be reasoned next is arbitrarily large. This challenges learning as it's hard to learn an approximation function that adapts well to arbitrary input length. Our experiments haven't achieved length generalization for problems with $R = \infty$ (see Table \ref{tab:experiment_property} and Figure \ref{fig:experiments}). 

\section{Related Work}
\label{sec.related}
This section reviews the related theoretical work. The empirical work is reviewed in \textit{Appendix}~\ref{appendix.related}, which mianly consists of evaluations of and improvements made to the reasoning capabilities of LLMs and Chain-of-Thought (CoT) and its variants.  

\cite{abbe2023generalization} studied out-of-distribution (OOD) generalisation of reasoning, in particular, an extreme case where part of the domain is entirely unseen during training, e.g., some value combinations are missing in training. It studies 
learning biases of different network architectures and activation functions. All resulting models may  
predict wrong on the OOD data.  {They also analyzed length generalization based on their proposed bias and used curriculum learning to improve the performance of the parity problem. We work on a different problem as we identify and prove conditions under which the length generalization can be solved and our results are independent of learning algorithms, activate functions, and architectures.} 


\cite{wies2023sub} proved that when sufficient intermediate steps (or CoT) are available, a neural network can efficiently learn any function in the \textbf{P} time complexity class. In addition, 
there exist functions in the \textbf{P} time complexity class that cannot be learned by any polynomial time learning algorithm without intermediate supervision (or CoT). Similarly, \cite{feng2023towards} showed why CoT works for mathematical and decision-making problems, especially those composite problems that can be decomposed into sub-problems, which the authors call dynamic programming problems (which is smilar to DAG in our paper). They first proved that it is impossible to predict the correct result directly without CoT, and then proved that CoT is capable of solving these problems. \cite{li2023dissecting} also attempted to demystify the mechanics underlying CoT. They showed that CoT can enable the model to identify each step (which they call \textit{filtering}) and then to work on the step (which they call \textit{in-context learning}) before moving to the next step in the CoT chain. 
\cite{malach2023auto} proved that that not only sophisticated auto-regressive next-token predictors like Transformer can be an universal learner, but also simple models such as linear next-token predictors, trained on CoT data. 
The paper also introduces the \textit{length complexity} to measure how many intermediate tokens are required to learn a function.

However, the theorems and proofs in these papers are based on the traditional i.i.d setting. 
Specifically, as discussed in Sec.~\ref{sec.overview}, these studies all under the given length/size $N$. Their statements are something like ``for $\forall\, N > 0$, given a dataset with training problems no longer than $N$, for any problem with length $N' \leq N$, it can be solved under some PAC upper bound.'' The primary limitation of their works is that either training problem length and testing problem length are the same, or the upper bound depends on the testing problem length. Our theory doesn't have this limitation and works on a more general scenario where training on smaller length problems and testing on larger length problems. Our statement is that ``for $\forall\, N > 0$, given a dataset with problems no longer than $N$, for any problem with arbitrary length $N'$, it can be solved if $|\mathbf{X}| < \infty$ and $R < \infty$.''

\section{Experiments}
\label{sec:experiment}

In previous sections, we have shown three key aspects for solving multi-step reasoning problems. 
\begin{enumerate}
    \item Given the structure of the problem in an MDP or a DAG, the causal function is well-learned only if the input space is finite i.e. $|\mathbf{X}| < \infty$, where the positive result is shown in Theorem \ref{thm: full_dynamic_function} and Theorem \ref{thm: full_causal_function} and the negative result is shown in Corollary \ref{thm: not_full_dynamic_function_inf_x} and Corollary \ref{thm: not_full_causal_function_inf_x}. 
    \item Given the structure of the problem and a well-learned causal function, the reasoning problem of arbitrary sizes can be recursively solved, which is shown in Theorem \ref{thm: recursive_markov_dp_solved} and Theorem \ref{thm: resursive_dag_solved}.
    \item Given unstructured sequence data (the realistic scenario), if $R < \infty$, we can find the elements to be calculated next to 
    solve the problem recursively via CoT and {to deal with length generalization}. The sufficient condition is $R = \sup \max_{v' \in (s \circ t)(v)} d(v', v) < \infty$, which is an important primitive property {that depends on the underlying DAG of the problem and the unstructured representation of the problem. {For the same problem, one representation may have $R = \infty$, but another representation may have $R < \infty$}.} 
    \vspace{-2mm}
\end{enumerate}

\begin{table}[h]
\vspace{-3mm}
\caption{Experimental settings 
} 
\vspace{1mm}
\centering
\scalebox{0.85}{
\begin{tabular}{c|cccccc}
\toprule
 & Train & Test 1 & Test 2 & Test 3 & Test 4 & Test 5 \\
\midrule
arctan & $r \in (1/2, 2)$ & $r \in (1/3, 3)$ & $r \in (1/4, 4)$ & $r \in (1/5, 5)$ & $r \in (1/6, 6)$ & $r \in (1/10, 10)$ \\
arithmetic in $F_7$  & $L \in [3, 20)$ & $L \in [3, 30)$ & $L \in [3, 40)$ & $L \in [3, 50)$ & $L \in [3, 60)$ & $L \in [3, 100)$ \\
1-line addition & $a, b \in [0, 10^6)$ & $a, b \in [0, 10^7)$ & $a, b \in [0, 10^8)$ & $a, b \in [0, 10^9)$ & $a, b \in [0, 10^{10})$ & $a, b \in [0, 10^{21})$ \\
3-line addition & $a, b \in [0, 10^6)$ & $a, b \in [0, 10^7)$ & $a, b \in [0, 10^8)$ & $a, b \in [0, 10^9)$ & $a, b \in [0, 10^{10})$ & $a, b \in [0, 10^{21})$ \\
1-line multiplication & $a, b \in [0, 5)$ & $a, b \in [0, 6)$ & $a, b \in [0, 7)$ & $a, b \in [0, 8)$ & $a, b \in [0, 9)$ & $a, b \in [0, 10)$ \\
\bottomrule

\end{tabular}
}
\label{tab:experiment_setting}
\vspace{-4mm}
\end{table}


Our experiments fall into three categories. The \textit{arctan} problem and the \textit{arithmetic in $F_7$} (the finite prime field with seven elements) problem verify the 1st aspect. The \textit{arithmetic} problem and the \textit{3-line addition} problem verify the 2nd aspect. The \textit{1-line addition} problem, \textit{3-line addition} problem and \textit{1-line multiplication} problem verify the 3rd aspect. 
{We list the detailed parameters of training and testing datasets of each problem in Table \ref{tab:experiment_setting} (explained below). The properties of each problem are listed in Table \ref{tab:experiment_property}, for ease of connecting our theory and experimental results. The accuracy results are reported in Figure \ref{fig:experiments}.} Some examples in each experiment are given in Table \ref{tab:examples} in \textit{Appendix}~\ref{appendix.examples}. The implementation details are also given in \textit{Appendix}~\ref{appendix.examples}. 

\vspace{-6mm}
\begin{wraptable}[10]{r}{3.8in}
\caption{Experimental primitive properties} 
\centering
\scalebox{0.9}{
\begin{tabular}{c|c|c}
\toprule
 & $|X|$ in Theorem \ref{thm: full_causal_function} & $R$ in Theorem \ref{thm: find_input_vertices_is_possible} \\
\midrule
arctan & $|X|=\infty$ & $R = 1$ \\
arithmetic in $F_7$ & $|X| = 13^5$ & $R=4$ \\
1-line addition & $|X| = 14^3$ & $R=\infty$  \\
3-line addition & $|X| = 14^9$ & $R=1$ \\
1-line multiplication & unknown & $R=\infty$ \\
\bottomrule
\end{tabular}
}
\label{tab:experiment_property}
\end{wraptable}
\vspace{+7mm}

The \textit{arctan} problem verifies a simple fact that the causal function may make mistakes when \textit{the input space is not finite}. The training data is sampled from an annulus of radius $r \in (1/2, 2)$. We test the performance on different annuluses, listed in Table \ref{tab:experiment_setting}. The performance decays as the annulus becomes larger, which satisfies Corollary \ref{thm: not_full_causal_function_inf_x}. 

\begin{wrapfigure}[9]{r}{3.5in}
\vspace{-2mm}
\caption{
Experimental results in accuracy. 
} 
\includegraphics[width=1.00\linewidth]{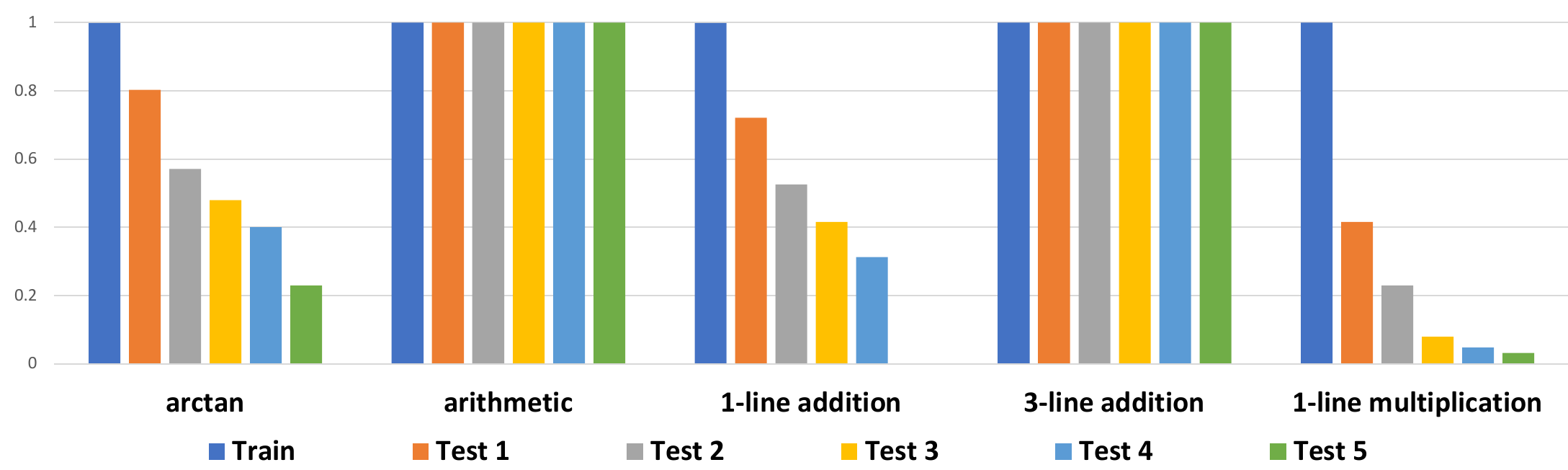}
\vspace{-7mm}
\label{fig:experiments}
\end{wrapfigure}

The \textit{arithmetic in $F_7$} problem verifies three things. First, the input space of the causal function is finite. For each calculation step, we need at most 5 elements as the input, e.g., $(1+2)$. Therefore, since $|\mathbf{X}| = |\{(,),+,-,\times, /,0,1,2,3,4,5,6\}|^5 < \infty$, by Theorem \ref{thm: full_causal_function}, the causal function can be well-learned. Second, it's obvious that when the DAG is known, we can simply apply the well-learned causal function to solve the problem of any length, which satisfies Theorem \ref{thm: resursive_dag_solved}. Third, it's important to ask how we acquire the underlying DAG. By Theorem \ref{thm: find_input_vertices_is_possible}, if $R < \infty$, it is possible to perfectly predict the elements that should be calculated next, which is like progressively following the underlying DAG, but there is no need to physically construct the DAG. In the \textit{arithmetic in $F_7$} problem, we have $R = 4$, because any combinations of elements that should be calculated next in a step is within a window of at most 5 elements, e.g., $(2+3)$. Finally, the problem satisfies all the positive theorems. The training and testing settings are listed in Table \ref{tab:experiment_setting}. The training data has at most 20 elements, i.e. $L \in [3, 20)$, and the testing data has at most 30, 40, 50, 60, 100 elements. The results in Figure \ref{fig:experiments} show this problem is perfectly solved. 

The \textit{1-line addition} problem and the \textit{3-line addition} problem consider the same DAG with different unstructured data for representation. They are trained on 5-digits additions, i.e. $a, b \in [0, 10^6)$, and tested on 5, 6, 7, 8, 9, 20-digits additions respectively. The \textit{1-line addition} problem fails to generalize beyond 5-digits, shown in Table \ref{fig:experiments}. The \textit{3-line addition} problem performs with $100\%$ accuracy on 5, 6, 7, 8, 9, 20-digits.
The two addition problems share the same DAG, but have different $R$'s due to their different representations. For \textit{1-line addition}, let us consider $x=\text{`}285+9805=?\text{'}$. The elements (or digits) to be calculated first are $x[2]=\text{`}5\text{'}$, $x[7]=\text{`}5\text{'}$ and $x[9]=\text{`}?\text{'}$. The maximal distance between them is $7$. This distance increases as number of digits increases. By definition, $R = \sup \max_{v' \in (s \circ t)(v)} d(v', v) = \infty$. 
But for 3-line addition, $x$ becomes
\begin{equation}
x = \left(\begin{aligned}
\ 285 \\
9805 \\
\ \ \ ?
\end{aligned}\right) = (\text{`}\ 285\text{'}, \text{`}9805\text{'}, \text{`}\ \ \ ?\text{'})^T.
\end{equation}
For each step, $x[i]$ only needs to consider its right neighbor $x[i+1]$. 
$x[3] = \left(\text{`}5\text{'}, \text{`}5\text{'}, \text{`}?\text{'}\right)^T$ is enough for calculation.
$x[2] = \left(\text{`}8\text{'}, \text{`}0\text{'}, \text{`}\ \text{'}\right)^T$ only needs to consider $x[3]$. 
The maximal distance of elements to be calculated next is always $1$, which doesn't depend on the number of digits. Therefore, $R = 1$. 

For the \textit{1-line multiplication} of two integers, 
we decompose multiplication into two stages. In the first stage, we transform multiplication into a summation of multiple integers. In the second stage, we solve the summation recursively. An example is shown in Table \ref{tab:examples}(d) in \textit{Appendix}~\ref{appendix.examples}. For the second stage, it's solvable by \textit{3-line addition} in principle. However, we have $R = \infty$ when dealing with the first stage. For instance, let input[k] = $\text{`}a \times b=\underbrace{a+\dots+a}_{k}+?\text{'}$. When $k < b - 1$, and output[k] = $\text{`}a \times b=\underbrace{a+\dots+a}_{k+1}+?\text{'}$. when $k = b - 1$, output[k] = $\text{`}a \times b=\underbrace{a+\dots+a}_{k+1}\text{'}$. In this example, Whether to add `$+?$' or to go to the second stage depends on $b$ and the number of existing $a$'s. Therefore, $b$ and all existing $a$'s are input elements in a reasoning step, where the maximal distance between them is arbitrary large as $b \rightarrow \infty$.
 
We have tried several other kinds of unstructured data for representing multiplication, including the traditional multi-line representation, but have not been able to design a good representation that solves the multiplication problem through CoT that can overcome the length generalization problem. 

\section{Discussions}
\label{sec: discussions}

In the discussion of the causal function, i.e. Theorem \ref{thm: full_dynamic_function}, Corollary, \ref{thm: not_full_dynamic_function}, Corollary \ref{thm: not_full_dynamic_function_inf_x}, Theorem \ref{thm: full_causal_function}, Corollary \ref{thm: not_full_causal_function} and Corollary \ref{thm: not_full_causal_function_inf_x}, the theories show only the existence of a positive/negative function. Thus, our theories are qualitative rather than quantitative and 
are not connected to specific learning algorithms or model structures. 
Quantitatively, in the context of neural networks, existed works~\cite{wies2023sub,feng2023towards} have shown the learnability of the CoT-based learning for reasoning under the PAC-learning framework. Our work focuses only on length generalization, which is learning paradigm and algorithm independent and clearly is applicable to neural networks. \cite{wies2023sub,feng2023towards} did not study the conditions for length generalization.

Our work considers reasoning problems that can be structured as a DAG. We don't know whether a reasoning problem (e.g., temporal and spatial reasoning) that cannot be represented as a DAG could be solved or not by CoT or the conditions under which it may be solvable to deal with length generalization. 

We found the maximal input elements distance $R$ to be an important quantity that decides whether a reasoning problem represented as a unstructured 
sequence could be learned to solve the length generalization problem. As demonstrated in the \textit{1-line addition} and \textit{3-lines addition} experiments, it is important to note the fact that different representations of the same reasoning problem may have different $R$'s, which may decide whether the problem is solvable or not solvable. Intuitively, the \textit{3-line addition} is more similar to how humans calculate a summation on a piece of 2-dimensional paper. We believe that it may not be rational to represent an inherently 2-dimensional task in one dimension like \textit{1-line addition}. Thus, using a suitable dimension to represent a reasoning problem may be important. 
An interesting question is whether all the reasoning problems could be represented by a high dimensional representation with $R < \infty$. In other words, it is unclear whether there exist any reasoning problem that has no CoT based representations with $R < \infty$. 
We leave these questions to our future work. 

\section*{Acknowledgments}

This work was initially inspired by a group discussion related to large language models at the \textit{Dagstuhl Seminar} on \textit{Deep Continual Learning} held from Mar 19 – Mar 24, 2023, for which Bing Liu was one of the organizers. 
Bing Liu would like to thank Muhan Zhang 
for helpful discussions. Bing Liu's work was supported in part by four National Science Foundation (NSF) grants (1910424, 1838770, 2225427,and 2229876) and a research contract from KDDI.



\bibliographystyle{alpha}
\bibliography{IAES2024}

\newpage
\appendix

\vspace{+3mm}
\subsection{Proofs - Causal Function}
\label{sec.dynamic.causal}
\vspace{+2mm}

We firstly provide Lemma \ref{lem: interpolation}, which is fundamental to our subsequent proofs. Lemma \ref{lem: interpolation} is similar to the \textit{universal approximation theorem} (UAT)~\cite{haykin1998neural,hassoun1995fundamentals,pinkus1999approximation}. We do not directly use the UAT as it is for neural networks but our results are not bounded to any specific learning paradigm or algorithm. But due to the UAT, our results are applicable to learning in neural networks. The UAT guarantees that $\sup_{x \in K}|| f(x) - g(x) || < \eta$, where $g(x)$ is the target function and $K$ is compact. Applying the UAT in our setting of Lemma \ref{lem: interpolation}, we have $\sup_{1 \leq i \leq n}|| f(x_i) - y_i || < \eta$, {\color{black}where $n$ is the number of training samples.} The UAT has a condition that $K$ is compact. In Lemma \ref{lem: interpolation}, since we have a stronger condition that $K = \{(x_i, y_i) |\, 1 \leq i \leq n\}$ is finite, we provide a stronger result that $\sup_{1 \leq i \leq n}|| f(x_i) - y_i || = 0$.

\begin{lemma}[A Simple Interpolating Function]
Let $(X, d_X)$ be a metric space. Let $(Y, ||\cdot||)$ be a Banach space. 
For any $D \subseteq \{(x, y) |\, x \in X, y\in Y\}$, if $x \neq x',\, \forall\, (x, y), (x',y') \in D$, then there exists a continuous approximation function $f: X \rightarrow Y$, s.t. $f(x) = y,\,\forall\, (x, y) \in D$.
\label{lem: interpolation}
\end{lemma}

\begin{proof}[Proof of Lemma \ref{lem: interpolation}]
Let 
$$
\epsilon = \min_{(x, y), (x', y') \in D} d_X(x, x').
$$
Define 
$$
K_\epsilon (x, x') = \frac{\epsilon}{d_X (x, x')}.
$$
Denote $D = \{(x_1, y_1), \dots, (x_n, y_n)\}$, let
\begin{equation}
    f(x) = \frac{\sum_{i = 1}^n y_i K_\epsilon(x, x_i)}{\sum_{i = 1}^n K_\epsilon(x, x_i)},\, x \in X \backslash \{x_1, \dots, x_n\}.
\label{eq: interpolate}
\end{equation}
For $\forall\, 1 \leq i_0 \leq n$  ($i_0$ is any integer), since $\lim_{x \rightarrow x'} K_\epsilon (x, x') = +\infty$, $\sup_{d_X(x, x')  > \epsilon} K_\epsilon (x, x') < 1$ and 
$$
f(x) = \frac{ y_{i_0} K_\epsilon(x, x_{i_0})}{\sum_{i = 1}^n K_\epsilon(x, x_i)} + \frac{\sum_{i \neq i_0} y_i K_\epsilon(x, x_i)}{\sum_{i = 1}^n K_\epsilon(x, x_i)},
$$  
it's obvious that $\forall\, \eta > 0$, $\exists\, \delta > 0$, s.t. $\forall\, d_X(x, x_{i_0}) < \delta$, $||f(x) - y_{i_0}|| < \eta$, which is $\lim_{x \rightarrow x_{i_0}} f(x) = y_{i_0}$. 
Therefore, we can define $f$ on $X$ and it's obvious that $f$ is a continuous approximation function. 
\end{proof}

~\\

\begin{proof}[\bf Proof of Theorem \ref{thm: full_dynamic_function}]
In the theorem presented in Section~\ref{sec:dp} in the main text, we wrote $D = Y \times X$ for simplicity. In detail, we have $D = \{((y, x), f(y, x))|\, (y, x) \in Y\times X\}$. 

By Lemma \ref{lem: interpolation}, let $Y \times X$ be $X$ and $Y$ be $Y$, there exists $\Hat{f}$ s.t. $\Hat{f}(y, x) = f(y, x),\,\forall\, (y, x) \in D$. Since $D = Y \times X$, the proof is done.
\end{proof}

~\\

\begin{proof}[\bf Proof of Corollary \ref{thm: not_full_dynamic_function}]
In the corollary presented in Section~\ref{sec:dp} in the main text, we wrote $D \neq Y \times X$ for simplicity. 
In detail, we have $D \neq \{((y, x), f(y, x))|\, (y, x) \in Y \times X\}$. 
For simplicity, denote $A = \{((y, x), f(y, x))|\, (y, x) \in Y \times X\}$. 

Since $D \neq A$, let $((y_0, x_0), f(y_0, x_0)) \in A \backslash D$. 
Since $|Y| > 1$, let $y^{err} \in Y$ s.t. $y^{err} \neq f(y_0, x_0)$. 
Let $$D^{err} = D \cup \{((y_0, x_0), y^{err})\}.$$

By Lemma \ref{lem: interpolation}, let $Y \times X$ be $X$ and $Y$ be $Y$, there exists $\Hat{f}$ s.t. $\Hat{f}(y, x) = y',\,\forall\, ((y, x), y') \in D^{err}$. 
Since $(y_0, x_0), y^{err}) \in D^{err}$, $\Hat{f}$ makes a wrong prediction as $$\Hat{f}(y_0, x_0) = y^{err} \neq f(y_0, x_0). $$
Since $D \subseteq D^{err}$, $\Hat{f}$ is correct on training dataset as $$\Hat{f}(y, x) = f(y, x),\,\forall\, ((y, x), f(y, x)) \in D.$$ 
\end{proof}

~\\

\begin{proof}[\bf Proof of Corollary \ref{thm: not_full_dynamic_function_inf_x}]
Denote $D \subseteq \{((y, x), f(y, x))|\, (y, x) \in Y \times X\}$. 
For simplicity, denote $A = \{((y, x), f(y, x))|\, (y, x) \in Y \times X\}$.

Since $|X| = \infty$, $|A| = \infty$. Since $|D| < \infty$, we know $D \neq A$ and $|A \backslash D| = \infty$. 
Let $((y_i, x_i), f(y_i, x_i)) \in A \backslash D,\,i=0,1\dots,m$. Since $|Y| > 1$, let $y_i^{err} \in Y$ s.t. $y^{err}_i \neq f(y_i, x_i)$. 
Let $$D^{err} = D \cup \{((y_i, x_i), y_i^{err}),\, i = 0, \dots, m)\}.$$

By Lemma \ref{lem: interpolation}, let $Y \times X$ be $X$ and $Y$ be $Y$, there exists $\Hat{f}$ s.t. $\Hat{f}(y, x) = y',\,\forall\, ((y, x), y') \in D^{err}$. 
Since $((y_i, x_i), y_i^{err}) \in D^{err}$, $\Hat{f}$ makes $m+1$ mistakes as $$\Hat{f}(y_i, x_i) = y_i^{err} \neq f(y_i, x_i),\,i = 0, 1, \dots, m. $$
Since $D \subseteq D^{err}$, $\Hat{f}$ is correct on training dataset as $$\Hat{f}(y, x) = f(y, x),\,\forall\, ((y, x), f(y, x)) \in D.$$ 
\end{proof}

~\\

\begin{proof}[\bf Proof of Theorem \ref{thm: full_causal_function}]
By Lemma \ref{lem: interpolation}, let $X^{\sup |s(v)|}$ be $X$ and $X$ be $Y$, there exists $\Hat{f}$ s.t. $\Hat{f}(s(v)) = v',\,\forall\, (v', s(v)) \in D$. 
Since $D = \{(f(s(v)), s(v)) |\, v \in V\}$, the proof is done. 
\end{proof}

~\\

\begin{proof}[\bf Proof of Corollary \ref{thm: not_full_causal_function}]
In the corollary presented in Section~\ref{sec: dag} in the main text, we wrote $D \neq X^{\sup |s(v)|}$ for simplicity. 
In detail, we have $D \neq \{(f(s(v)), s(v)) |\, v \in V\}$. 
For simplicity, denote $A = \{(f(s(v)), s(v)) |\, v \in V\}$. 

Since $D \neq A$, let $(f(s(v_0)), s(v_0)) \in A \backslash D$. 
Since $|\{f(s(v)|\, s(v) \in X^{\sup |s(v)|})\}| > 1$, let $v^{err} \in \{f(s(v)|\, s(v) \in X^{\sup |s(v)|})\}$ s.t. $v^{err} \neq f(s(v_0))$. 
Let $$D^{err} = D \cup \{(v^{err}, s(v_0))\}.$$

By Lemma \ref{lem: interpolation}, let $X^{\sup |s(v)|}$ be $X$ and $X$ be $Y$, there exists $\Hat{f}$ s.t. $\Hat{f}(s(v)) = v',\,\forall\, (v', s(v)) \in D^{err}$. 
Since $(v^{err}, s(v_0)) \in D^{err}$, $\Hat{f}$ makes a wrong prediction as $$\Hat{f}(s(v_0)) = v^{err} \neq f(s(v_0)). $$
Since $D \subseteq D^{err}$, $\Hat{f}$ is correct on training dataset as $$\Hat{f}(s(v)) = f(s(v)),\,\forall\, (f(s(v)), s(v)) \in D.$$ 
\end{proof}
 
 ~\\

\begin{proof}[\bf Proof of Corollary \ref{thm: not_full_causal_function_inf_x}]
Denote $D \subseteq \{(f(s(v)), s(v)) |\, v \in V\}$. 
For simplicity, denote $A = \{(f(s(v)), s(v)) |\, v \in V\}$. 

Since $\max(|X|, \sup |s(v)|) = \infty$, $|X^{\sup |s(v)|}| = \infty$, $|A| = \infty$. Since $|D| < \infty$, we know $D \neq A$ and $|A \backslash D| = \infty$. 
Let $(f(s(v_i), v_i)) \in A \backslash D,\,i=0,1\dots,m$. Since $|\{f(s(v)|\, s(v) \in X^{\sup |s(v)|})\}| > 1$, let $v_i^{err} \in \{f(s(v)|\, s(v) \in X^{\sup |s(v)|})\}$ s.t. $v^{err}_i \neq f(s(v_i))$. 
Let $$D^{err} = D \cup \{(f(s(v_i), v_i)),\, i = 0, \dots, m)\}.$$

By Lemma \ref{lem: interpolation}, let $X^{\sup |s(v)|}$ be $X$ and $X$ be $Y$, there exists $\Hat{f}$ s.t. $\Hat{f}(s(v)) = v',\,\forall\, (v', s(v)) \in D^{err}$. 
Since $(v_i^{err}, s(v_i)) \in D^{err}$, $\Hat{f}$ makes $m + 1$ mistakes as $$\Hat{f}(s(v_i)) = v_i^{err} \neq f(s(v_i)), \,i = 0, 1, \dots, m. $$
Since $D \subseteq D^{err}$, $\Hat{f}$ is correct on training dataset as $$\Hat{f}(s(v)) = f(s(v)),\,\forall\, (f(s(v)), s(v)) \in D.$$ 
\end{proof}

\subsection{Proofs - Recursive Formula}
\label{sec.recursive.direct}
\vspace{+2mm}

\begin{proof}[\bf Proof of Theorem \ref{thm: recursive_markov_dp_solved}]
By Theorem \ref{thm: full_dynamic_function}, there exists $\Hat{f}$ s.t. $\Hat{f}(y, x) = f(y, x),\,\forall\, (y, x) \in Y \times X$. 

\noindent
Since $\tau$ is a Markov dynamic process, by the definition in Eq. \eqref{eq: def_dp}, we have
$$
\tau_f( x_{[0: \infty]}) = (y_1, y_2, \dots),\ \text{where}\  \left\{
\begin{aligned}
    &y_1 = f(y_0, x_0), \\
    &\dots \\
    &y_{n + 1} = f(y_n, x_{n}). \\
\end{aligned}
\right.
$$
By induction, we have
$$
\begin{aligned}
    y_1 = f(y_0, x_0) &= \Hat{f}(y_0, x_0) = \Hat{y}_1, \\
    &\dots \\
    y_{n + 1} = f(y_n, x_{n}) &= \Hat{f}(\Hat{y}_n, x_n) = \Hat{y}_{n + 1}. \\
\end{aligned}
$$
\end{proof}

~\\ 






\begin{proof}[\bf Proof of Theorem \ref{thm: resursive_dag_solved}]
By Theorem \ref{thm: full_causal_function}, there exists $\Hat{f}$, s.t. $\Hat{f}(s(v)) = f(s(v)),\,\forall\, (f(s(v)), s(v)) \in \{(f(s(v)), s(v)) |\, v \in V\}$. 

By the definition in Eq. \eqref{eq: def_dag} in Section~\ref{sec: dag}, for an arbitrary topological ordering of the graph $G$, we have 
$$
G_f(\{v_i, i \leq |G| |\, d(v_i) = 0\}) = (v_1, \dots, v_{|G|}), \text{where}\ 
\left\{
\begin{aligned}
&v_1 = f(s(v_1)), \\
&\dots \\
&v_n = f(s(v_n)).
\end{aligned}
\right.
$$

Given the graph $G$ and the topological ordering, by induction, we have
$$
\begin{aligned}
v_1 = f(s(v_1)) &= \Hat{f}(s(v_1)) = \Hat{v}_1, \\
&\dots \\
v_n = f(s(v_n)) &= \Hat{f}(s(\Hat{v}_n)) = \Hat{v}_n. 
\end{aligned}
$$
\end{proof}

\subsection{Proof - Maximal Input Element Distance of a Reasoning step}
\label{sec.mied}

\begin{wrapfigure}[9]{r}{2.8in}
\vspace{-4mm}
\caption{
Two examples of the \textit{ko} problem. 
} 
\includegraphics[width=0.8\linewidth]{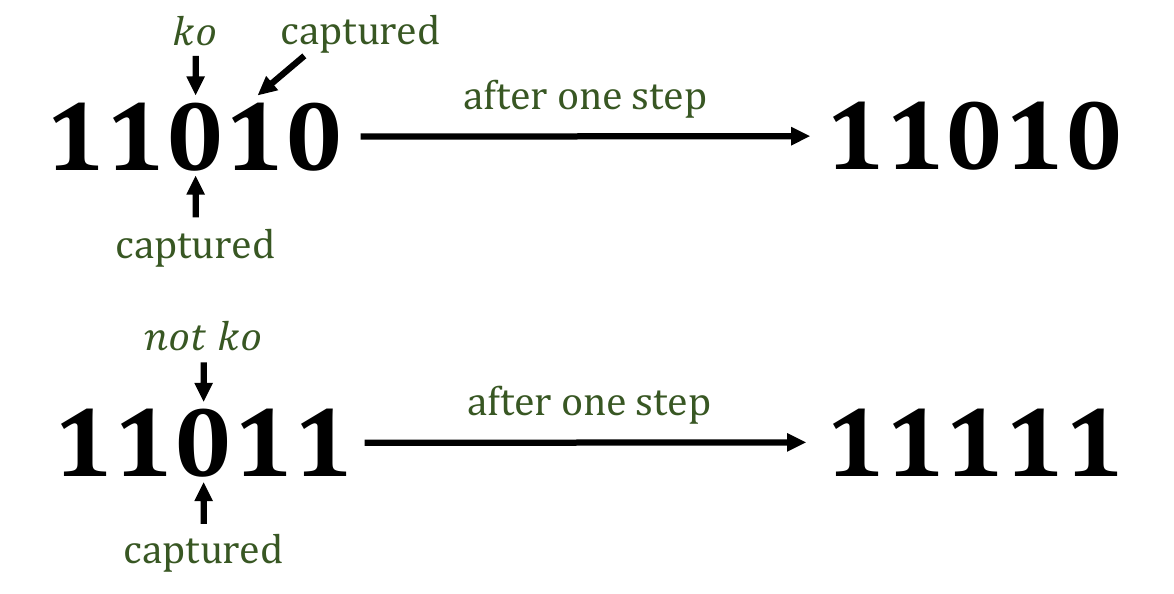}
\label{fig:example_ko}
\end{wrapfigure}

Before presenting the proof for Theorem~\ref{thm: find_input_vertices_is_possible}, we first provide an example to show that a $4R$-length window is not enough to decide which elements should be used in the next reasoning or calculation step. 

Let's consider a simple one-dimensional \textit{ko} problem. Let $s = \dots 0 0 1 0 0 1 1 1 1 \dots$ be a sequence of $0$'s and $1$'s. We say $s_i$ is captured if $s_{i} \neq s_{i-1},\,s_{i} \neq s_{i+1}$. We say $s_i$ is not a ko if both $s_{i-1}$ and $s_{i+1}$ are not captured. We say $s_i$ is a ko if $s_i$ is captured and at least one of $s_{i-1}$ and $s_{i+1}$ is captured. When some $s_i$ is captured and is not a ko, {we set $s_i = 1 - s_i$.} We continue the process until $s$ won't change, which is the final settlement of the problem.
In this problem, if $s_i$ is captured and is not a ko, {it should be acted upon next (changing $s_i$'s value).} The neighbors $s_{i-1}$ and $s_{i+1}$ are enough to decide the value of $s_i$, i.e. $\max_{v' \in (s \circ t)(s_i)} d(v', s_i) = \max(d(s_{i-1}, s_i), d(s_{i+1}, s_i)) = 1$. Therefore, $R = 1$ in this problem. 

However, a $4R$-length window is not enough to decide whether $s_i$ should be reasoned or acted upon next. For instance, let $s = 1 1 0 1 0$. In this problem, $s_{[2]} = 0$ is captured and is a ko, so it cannot be changed. Therefore, when we only consider $4$ elements $s_{[0:4]} = 1 1 0 1$, we will receive the ground truth label $y_{[0:4]}= 0 0 0 0$, where $0$ indicates the element shouldn't be changed and $1$ indicates the element should be changed. However, when we consider a problem $s' = 1 1 0 1 1$, $s_{[2]}' = 0$ is captured and is not a ko, so it should be changed. Therefore, when we only consider $4$ elements $s_{[0:4]}' = 1 1 0 1$, the ground truth label is $y_{[0:4]}'= 0 0 1 0$. In these two problems, they share an identical $4R$-length window $1 1 0 1$, but the labels of this window are different, which is obviously ambiguous. 

\vspace{+2mm}
\begin{proof}[\bf Proof of Theorem \ref{thm: find_input_vertices_is_possible}]
Denote the training dataset as
$$
D \subseteq \{(\Tilde{g}(s), s) | s = s_0 s_1 \dots s_{4R}, s = g^{-1}(G)\}, 
$$
where $\Tilde{g}(s) = 1_{0}(s_0) 1_{1}(s_1) \dots 1_{4R}(s_{4R})$ is a $(4R+1)$-length binary sequence that $1_i (s_i)$ indicates whether $s_i$ should be reasoned next. 

By Lemma \ref{lem: interpolation}, let $X^{4R + 1}$ to be $X$ and $2^{4R+1}$ to be $Y$, there exists $\Hat{f}$ s.t. $\Hat{f}(s) = I,\,\forall\, (I, s(v)) \in D$. 
Since $D = \{(\Tilde{g}(s), s) | s = s_0 s_1 \dots s_{4R}, s = g^{-1}(G)\}$, the proof is done. 
\end{proof}

\vspace{+3mm}

\begin{table}[h]
\fontsize{9pt}{9pt}\selectfont 
\caption{
Experimental data examples 
} 
\centering
\begin{minipage}[l]{0.35\textwidth}
\vspace{-5mm}
\subfigure[arctan]{
\scalebox{1.1}{
\begin{tabular}{l|l}
\toprule
Input[0]:     & a, b \\
Output[0]:      & arctan(a/b) \\
\bottomrule
\end{tabular}
}
}
~\\~\\~\\~\\~\\~\\~\\~\\~\\~\\
\subfigure[arithmetic in prime field $F_7$]{
\scalebox{1.1}{
\begin{tabular}{l|l}
\toprule
Input[0]:      &  (0+4-(2-3*6))*(4+0) \\
Output[0]:      &  (\ \,4\ \ -(2-\,\,4\,\ ))*\ \ 4 \\
\midrule
Input[1]:      & (4-(2-4))*4  \\
Output[1]:      & (4-\ \ 5\,\ \ )*4 \\
\midrule
Input[2]:      & (4-5)*4  \\
Output[2]:      & \ \ \,6\ \ *4 \\
\midrule
Input[3]:      &  6*4 \\
Output[3]:      & \ \,3 \\
\bottomrule
\end{tabular}
}
}
\end{minipage}
\begin{minipage}[l]{0.35\textwidth}
\scalebox{1.1}{
\subfigure[1-line addition]{
\begin{tabular}{l|l}
\toprule
Input[0]:  &  285+9805=? \\
Output[0]: &  285+9805=c0 \\
\midrule
Input[1]:  &  285+9805=c0 \\ 
Output[1]: & 285+9805=?90 \\
\midrule
Input[2]:  &  285+9805=?90 \\
Output[2]: & 285+9805=c090 \\
\midrule
Input[3]:  &  285+9805=c090 \\ 
Output[3]: & 285+9805=10090 \\
\bottomrule
\end{tabular}
}
}
\subfigure[1-line multiplication]{
\scalebox{1.1}{
\begin{tabular}{l|l}
\toprule
Input[0]:  &  1*3=? \\
Output[0]: & 1*3=1+? \\ 
\midrule
Input[1]:  &  1*3=1+? \\
Output[1]: & 1*3=1+1+? \\ 
\midrule
Input[2]:  &  1*3=1+1+? \\
Output[2]: & 1*3=1+1+1 \\ 
\midrule
Input[3]:  &  1*3=1+1+1 \\
Output[3]: & 1*3=2+1 \\
\midrule
Input[4]:  &  1*3=2+1 \\
Output[4]: & 1*3=3 \\
\bottomrule
\end{tabular}
}
}
\end{minipage}
\begin{minipage}[r]{0.28\textwidth}
\vspace{4mm}
\subfigure[3-line addition]{
\scalebox{1.1}{
\begin{tabular}{l|r}
\toprule
Input[0]:  &   89283 \\
  &     3360 \\
  &          ? \\
Output[0]:   &     ?3 \\
\midrule
Input[1]:   &  89283 \\
   &    3360 \\
   &        ?3 \\
Output[1]:	&  c43 \\
\midrule
Input[2]:   &  89283 \\ 
   &    3360 \\
   &      c43 \\
Output[2]:  &  ?643 \\
\midrule
Input[3]:   &  89283 \\
   &    3360 \\ 
  &    ?643 \\
Output[3]:  & c2643 \\
\midrule
Input[4]:   &  89283 \\
   &    3360 \\
   &  c2643 \\
Output[4]:  & 92643 \\
\bottomrule
\end{tabular}
}
}
\end{minipage}
\label{tab:examples}
\end{table}

\subsection{Experimental Data and Implementation Details}
\label{appendix.examples}

The \textit{arithmetic in $F_7$} problem is defined in the prime field $F_7$. The \textit{1-line addition} problem and the \textit{3-line addition} problem use `?' to represent $0$ being carried from the right and `c' to represent $1$ being carried from the right. In all problems, * is equivalent to $\times$. We use * instead of $\times$ for ease of aligning chars. 

All the problems are trained with $50k$ batches. Each batch contains $256$ CoT steps. For each problem, we first randomly generate a question and then its detailed CoT steps. Each CoT step is a pair (Input[i], Output[i]), as shown in Table \ref{tab:examples}. We put the CoT steps into a batch until the batch size reaches $256$. 

When testing the performance after training, we test $6$ different datasets for each problem, which is shown in Table \ref{tab:experiment_setting}. {Each testing dataset is independently generated with $1k$ questions. We solve each question by CoT using the trained model.}
For the \textit{arctan} problem, since it only has one step, we only inference in one step. 
{For the \textit{arithmetic} problem, we stop the CoT output generation in a step when (i) the output of the step has only one token/element, or (ii) the output of the step is identical to the input of the step. }
For \textit{1-line addition} and \textit{3-line addition}, we stop the CoT output generation if (i) `?' and `c' aren't shown in the output, or (ii) the number of CoT steps is greater than the number of digits.
For the \textit{1-line multiplication} problem, we stop the CoT output generation if (i) `+' is not shown in the output, or (ii) the number of CoT steps is greater than the multipliers.

For the \textit{arctan} problem, the predicted value is considered correct if the absolute error is smaller than $0.01$. For the other problems, the final output is considered correct only when it is identical to the ground truth. 
For the dataset of each problem, the accuracy is the number of correctly answered questions divided by the total number of questions. 

The model of the \textit{arctan} problem has 3 fully connected layers. 
The model of the \textit{arithmetic} problem, the \textit{1-line addition} problem, the \textit{3-lines addition} problem and the \textit{1-line multiplication} problem has 3 Transformer encoders with relative position embedding. 
{The optimizer is Adam and the learning rate is $0.0001$. The training data for each task contains $12.8$M CoT steps and was trained for $1$ epoch.}


\subsection{Related Empirical Work}
\label{appendix.related}
\textbf{Evaluations and limitations of LLM reasoning}. Continuing with the discussion about evaluations of the reasoning capabilities of LLMs in Section~\ref{sec-intro}, we present a more extensive literature survey here. Note that Sections~\ref{sec-intro} and \ref{sec.overview} have discussed empirical works about length generalization~\cite{anil2022exploring,dziri2023faith,zhang2022unveiling}. In these papers, the authors also tried to mitigate the problem through improved training~\cite{anil2022exploring}, improved promoting and fine-tuning of LLMs~\cite{zhang2022unveiling}, and curriculum learning~\cite{abbe2023generalization}. However, none of them studied the length generalization problem theoretically as we do. Below, we focus on surveying other empirical works. Many of them identified limitations of LLMs in solving different reasoning problems, but few have characterized the limitations in a formal manner to facilitate theoretical investigation.

\cite{meadows2023symbolic} created a dataset specifically for mathematical reasoning that can be perturbed. They showed that perturbations of the tasks heavily affect the results, reducing F1 score from 97\% to 17\%, which suggests that inference is likely to be dominated by surface-level patterns unrelated to the deeper understanding of the mathematical operators. However, this evaluation was done using only BERT~\cite{devlin2018bert} based models, but not on more recent LLMs like ChatGPT and GPT4. \cite{wu2023reasoning} used ``counterfactual'' tasks that deviate from the standard reasoning tasks to evaluate LLMs. It was found that the performance degrades substantially compared to the default conditions, which again suggests that while LLMs can perform reasoning to some extend, they often rely on narrow, non-transferable procedures or surface patterns for task-solving. A counterfactual based evaluation was also done in \citep{li2023counterfactual}, which reached the same conclusion.   

\cite{liu2023evaluating} evaluated ChatGPT and GPT-4 on logical reasoning. The results showed that they do relatively well on well-known public domain datasets, but their performances drop substantially when newly released and out-of-distribution datasets are used. \cite{xu2023large} too evaluated LLMs using logical reasoning  (deductive, inductive, abductive and mixed-form reasoning) and gave pros and cons of LLMs. 
\cite{she2023scone} created a dataset for reasoning involving negations and evaluated LLMs and showed poor results. \cite{ando2023evaluating} created a dataset, originally designed for psychological experiments to assess human logical abilities in syllogistic reasoning. 
The authors examined three types of biases observed in human syllogistic reasoning: \textit{belief biases}, \textit{conversion errors}, and \textit{atmosphere effects}. The evaluation on LLMs showed that they struggle with problems involving these biases too. \cite{tan2023towards} created a dataset to evaluate LLMs on temporal reasoning and showed some weaknesses of LLMs. They then proposed an approach to improve the results.   

\textbf{Chain of thoughts (CoT) and variants}. Earlier prompting for solving reasoning problems using LLMs only states the question and the answer. They found that these two pieces of information are insufficient for LLMs to learn to perform effective reasoning, which we have proved. Then \textit{chain of thought} (CoT) prompting~\citep{wei2022chain} was proposed to improve the situation. CoT basically contains
the detailed intermediate reasoning steps between the question and the answer for fine-tuning the LLMs, which significantly enhance LLMs' reasoning capabilities~\cite{chung2022scaling,hsieh2023distilling,mukherjee2023orca,fu2023chain}. \cite{saparov2022language} created a synthetic dataset generated based on first-order logic. They then parsed the generated CoT into symbolic proofs for formal analysis. It was shown that LLMs are capable of reasoning. The success of CoT has encouraged researchers to refine the technique and also propose variations of the technique. 

For example,~\cite{chen2023measuring} proposed a metric to measure the effectiveness of CoT and a technique to improve CoT for vision-language models. \cite{wang2023making} studied using multiple reasoning paths and positive and negative answers to improve CoT reasoning. 
\cite{zhang2023cumulative} proposed cumulative reasoning, which employs LLMs in a cumulative and iterative manner to emulate the human thought process. \cite{qi2023art} proposed a divide-and-conquer algorithm that simulates the self-questioning and recursive thinking process of humans to improve CoT.~\cite{wang2023learning} investigated how to incorporate into relatively small LMs the capabilities of multi-step reasoning and CoT.
\cite{wang2022towards} found that even logically invalid CoT also helps reasoning. This was confirmed in~\citep{schaeffer2023invalid}. To deal with unsound inferences, \cite{poesia2023certified} introduced a class of external tools for LLMs called guides that use states and incremental constraints to guide the generation in reasoning. A related work on using external tools was done in~\cite{xu2023rewoo}.
\cite{wang2022self} improved CoT using multiple-paths and consistency check. 
\cite{ling2023deductive} studied the verification of CoT. \cite{stolfo2023understanding} identified part of an LLM responsible for reasoning. In a different direction,~\citep{yang2022chain} argued that the prevailing approach to CoT prompt selection through trial and error is unsatisfactory. They then proposed a principled approach for multi-domain LLM CoT prompt selection.

Several researchers also broadened the CoT method and proposed the neural symbolic \textit{code prompting}~\citep{hu2023code},
\textit{program of thoughts}~\citep{chen2022program,cheng2023binding}, \textit{tree-of-thoughts}~\citep{yao2023tree,long2023large}, \textit{tree-of-mixed-thoughts}~\citep{hu2023tree}, \textit{tree of uncertain thoughts}~\citep{mo2023tree}, \textit{hypergraph-of-thoughts}~\citep{yao2023thinking}, \textit{recursion of thoughts}~\citep{lee2023recursion}, \textit{chain of knowledge}~\citep{wang2023boosting}, \textit{chain of simultaneous thoughts}~\citep{shao2022chaining}, \textit{graph-of-thoughts}~\citep{yao2023beyond} and \textit{faithful chain of thoughts}~\citep{lyu2023faithful}. 
Further, \cite{bi2023program} proposed a complexity measure and chose the optimal complexity to improve the \textit{program of thoughts}~\citep{chen2022program}. \cite{wang2023exploring} proposed a method to improve the generation of equations from natural language questions as the intermediate step to answer the original question. \cite{gao2023pal} combined CoT and Program-Aided Language Models (PAL) for improved reasoning.

\end{document}